\documentclass{llncs}

\usepackage{makeidx}

\usepackage{times}
\usepackage{epsfig}
\usepackage{graphicx}
\usepackage{amsmath}
\usepackage{amssymb}
\usepackage[breaklinks=true,bookmarks=false]{hyperref}
\usepackage{color}

\newcommand{\rbt}{RBST} 
\newcommand{\rbtfull}{Random Binary Search Trees}

\newcommand{\aqt}{average query time}

\newcommand{\amc}{average memory consumption}
 
\newcommand{\prten}{Precision@10}
 
\newcommand{\recallten}{Recall@10}

\newcommand{\notr}{$N_{tree}$} 
\newcommand{\md}{$D$} 
 
\newcommand{\mnot}{$N_{test}$}

\begin{document}

\mainmatter              

\title{\rbtfull{} for Approximate Nearest Neighbour Search in Binary Space}
\titlerunning{\rbtfull{}}  

\author{Micha\l{} Komorowski \and Tomasz Trzci\'{n}ski}
\authorrunning{M. Komorowski and T. Trzci\'{n}ski} 
\tocauthor{}
\institute{Institute of Computer Science,\\
Warsaw University of Technology,\\
Nowowiejska 15/19, Warsaw 00-665, Poland\\
\email{michalkomorowski1984@gmail.com, t.trzcinski@ii.pw.edu.pl}}

\maketitle 

\begin{abstract}
Approximate nearest neighbour (ANN) search is one of the most important problems in computer science fields such as data mining or computer vision. In this paper, we focus on ANN for high-dimensional binary vectors and we propose a simple yet powerful search method that uses \rbtfull{} (\rbt{}). We apply our method to a dataset of 1.25M binary local feature descriptors obtained from a real-life image-based localisation system provided by Google as a part of Project Tango \cite{GoogleTango}. An extensive evaluation of our method against the state-of-the-art variations of Locality Sensitive Hashing (LSH), namely Uniform LSH and Multi-probe LSH, shows the superiority of our method in terms of retrieval precision with performance boost of over 20\%.
\end{abstract}

\section{Introduction}

The goal of nearest neighbour search is to find vectors from a database that lie close to a query vector. This is a common use case in disciplines such as computer vision~\cite{Torralba08} or data mining~\cite{Shakhnarovich03}. However, often finding the exact nearest neighbour is costly while retrieving approximate neighbours is sufficient. Therefore several successful solutions in the area of Approximate Nearest Neighbour Search (ANN) have been proposed and among them the two most prominent ones are hierarchical structure (tree) based methods \cite{Bentley75,Fukunaga75} and hashing based methods \cite{Gionis99,Weiss09}.

One of the typical computer vision tasks where ANN search is used due to prohibitive amounts of data points is image-based localisation~\cite{feng2016fast,sattler2011fast}. ANN search is typically used in this context to find similarities between local feature descriptors extracted from different images. The majority of works on ANN focus on descriptors that are vectors of real numbers \cite{Fukunaga75,Liu2004,Nister2006,Silpa2008}. However, extraction of real-valued descriptors is time consuming so they are often substituted with binary descriptors when real-time performance is required. At the same time methods suitable for real-valued descriptors do not seem to work equally well when applied to binary ones~\cite{trzcinski2012thick}.

In this paper, we propose ANN search method that uses \rbtfull{} (\rbt{}) to find similar vectors within a database of binary vectors. As a use case of our method we take image-based localisation problem and we evaluate our method on a real world dataset of over 1 million binary local feature descriptors obtained within the frames of Google Project Tango \cite{GoogleTango} collaboration. Our ANN search method outperforms the state of the art in terms of retrieval accuracy, while providing similar recall and memory consumption.

Several other types of trees have been proposed in the literature for indexing of binary descriptors e.g.: k-means trees, kd-trees, or vantage-points trees \cite{kumar2008good}. However, their application to binary descriptors leads to severe performance drops, as indicated in~\cite{trzcinski2012thick}. Therefore we compare our proposed \rbtfull{} method with Local Sensitivity Hashing method~\cite{Gionis99} and its further modifications: Uniform LSH~\cite{trzcinski2012thick} and Multi-probe LSH~\cite{Lv07}.

\section{\rbtfull{}}
\label{lab:random_binary_trees}

In this section, we propose a simple yet powerful ANN method for indexing and searching a database of binary descriptors. We draw the inspiration for the method from standard Binary Search Trees (BST) \cite{Cormen}. These structures are well designed for speeding up search process and building up on their success, we propose a modified version of them, called \rbtfull{}. Our proposed \rbt{} differ from standard Binary Search Trees in the following aspects. 

Firstly, all paths from the root node to the leafs in our \rbt{} have the same length. Secondly, the leafs of our \rbt{} are used to store binary descriptors. The most important difference, however, is the fact that the nodes of our trees store a bit mask. It specifies which bit of a binary descriptor needs to be checked in order to decide if a given descriptor should be assigned to the left or to the right branch of a given node during indexing and search. Thanks to this setup \rbt{} are extremely fast as no distances must be calculated in order to create and search them - the fast binary operation AND is used instead. 

In the indexing stage, we use one or more \rbtfull{} to store the information about binary descriptors from our database.  Each descriptor from the database traverses the tree from the root towards the leaves. While traversing the tree, the descriptor is assigned to left or right branch based on the output of the binary AND operation on the descriptor and the bit mask of the node. In the querying stage, we use those constructed trees to search for candidate nearest neighbours by traversing the trees with a query descriptor and retrieving candidates per each tree. The final set of candidates is returned as a union of candidates across the trees. In the last stage of search, candidate descriptors are sorted based on their Hamming distance to the query descriptor. We then retrieve $N$ descriptors with the smallest distance.

Our \rbtfull{} algorithm is controlled by four parameters: $N$ equals to number of approximate nearest neighbours retrieved with default $N = 10$, \notr{} defines the number of \rbt{} to be created, \md{} specifies the maximum depth of a tree and \mnot{} defines how many dimensions of a binary descriptor can be checked in a single tree. Although each node can check only one dimension, this parameter allows us to randomly subsample the space of binary dimensions across different binary trees and increases robustness of our method.
    
The randomness of our \rbt{} stems from the fact that bits masks for nodes are selected randomly from a given set of bits. A similar idea is used in \cite{lepetit2006keypoint}. However, the trees proposed in~\cite{lepetit2006keypoint} were not used to index binary descriptors, but to classify keypoints. A related method can also be found in \cite{feng2016fast} where trees are generated in supervised way using a stability metric. We evaluated application of this approach to our \rbt{}, however, in our experiments trees proposed in \cite{feng2016fast} were up to 3 orders of magnitude slower than our \rbtfull{}. Our proposed \rbt{} may also look similar to Randomised Binary Search Trees \cite{seidel1996randomized}. However, there are few differences. In comparison to our \rbt{} data structure proposed in \cite{seidel1996randomized} associates a priority with every inserted key, use rotations to balance a tree and does not store list of keys (in our case descriptors) in leafs.

\subsection{Bits Selection Metrics and Hash Codes}

We also used the following bit metrics to weight the probability of a given bit to be selected for a mask in the nodes: Shannon entropy of a bit, its conditional entropy and its empirical stability. We define the empirical stability metric as a number of descriptors representing the same 3D point with equal value of a given bit to a total number of descriptors of the same 3D point. After calculating those bit metrics, we used their distribution as bias in the selection of a bit mask for each node. Bits with the higher values of bit metrics are used more often to generate \rbt{}. In order to limit memory consumption, we also used hash codes of binary descriptors, instead of the raw vectors. For hashing the descriptors we used Semi-Supervised Hashing method \cite{WangSSH2012} with various hash code lengths (32, 64, 128, 256 bits). Although in some cases bit metrics or hash codes increased the performance of our method, the improvement was rather negligible and, therefore, in the remainder of this paper, we rely on random bit selection for the node masks.
\vspace{-0.05cm}
\section{Evaluation}

In this section, we evaluate the accuracy and efficiency of our \rbt{}  method and compare it with the state of the art.  To increase robustness of our evaluation, we run our experiments 10 times, each time on a different subset of 100K descriptors extracted from dataset of 1.26M 512-dimensional binary FREAK descriptors~\cite{Alahi12}. This dataset was obtained from Google Project Tango~\cite{GoogleTango} collaboration and was generated using state-of-the-art 3D reconstruction methods. As evaluation metrics, we use Precision@N defined as number of correctly retrieved nearest neighbours within the first $N$ descriptors retrieved. Similarly, we compute Recall@N defined as the ratio of retrieved nearest neighbours describing the same 3D point within $N$ returned descriptors versus all descriptors describing given 3D point. We also measure querying time and average the results over 10 runs. All experiments are run using a server with 32 GB of RAM and Intel(R) Xeon(R) 2.60GHz CPU. 

\subsection{Initial Experiments}
\label{lab:crt_finding_the_winning_configuration}

To validate our method and verify the appropriate range of parameters, we first run an initial set of experiments with the following set of parameters: \notr{}~$= \{1, 3, 6, 9, 12\}$, \md{}~$=\{20, 30, 40, 50\}$ and \mnot{}~$=\{64, 128, 256, 512\}$. Based on obtained results, we defined a default set of parameters to be evaluated against the state of the art in the next sections, as they give a good balance between the precision, the recall and the average query time: \md{} $ = \{30, 40, 50\}$ and \mnot{}$ = 256$. 

We also discovered the following trends. Firstly, the higher value of \notr{} the higher \prten{}, at the expense of the \aqt{}. The dependence between \notr{} and the \aqt{}, assuming other parameters remain unchanged, is quasi-linear.  Secondly, the lower value of \md{}, the higher \aqt{}. Shallower trees have leafs with higher number of descriptors and since the last step of search includes sorting candidate vectors, the more candidates we retrieve, the longer the sorting. The highest precision can be obtained for the trees with the highest depth, for which majority of leafs contain not more than a few nodes but at the cost of the recall. As to \mnot{}, the higher value of this parameter the smaller \aqt{} (even 2 times or more), because descriptors are spread across higher number of leafs. This, in turn, results from a fact that more bits are taken into account while generating trees. 

\subsection{Comparison with the State of The Art}
\begin{figure}[t]
    \centering
    \includegraphics[width=\textwidth,height=7cm]{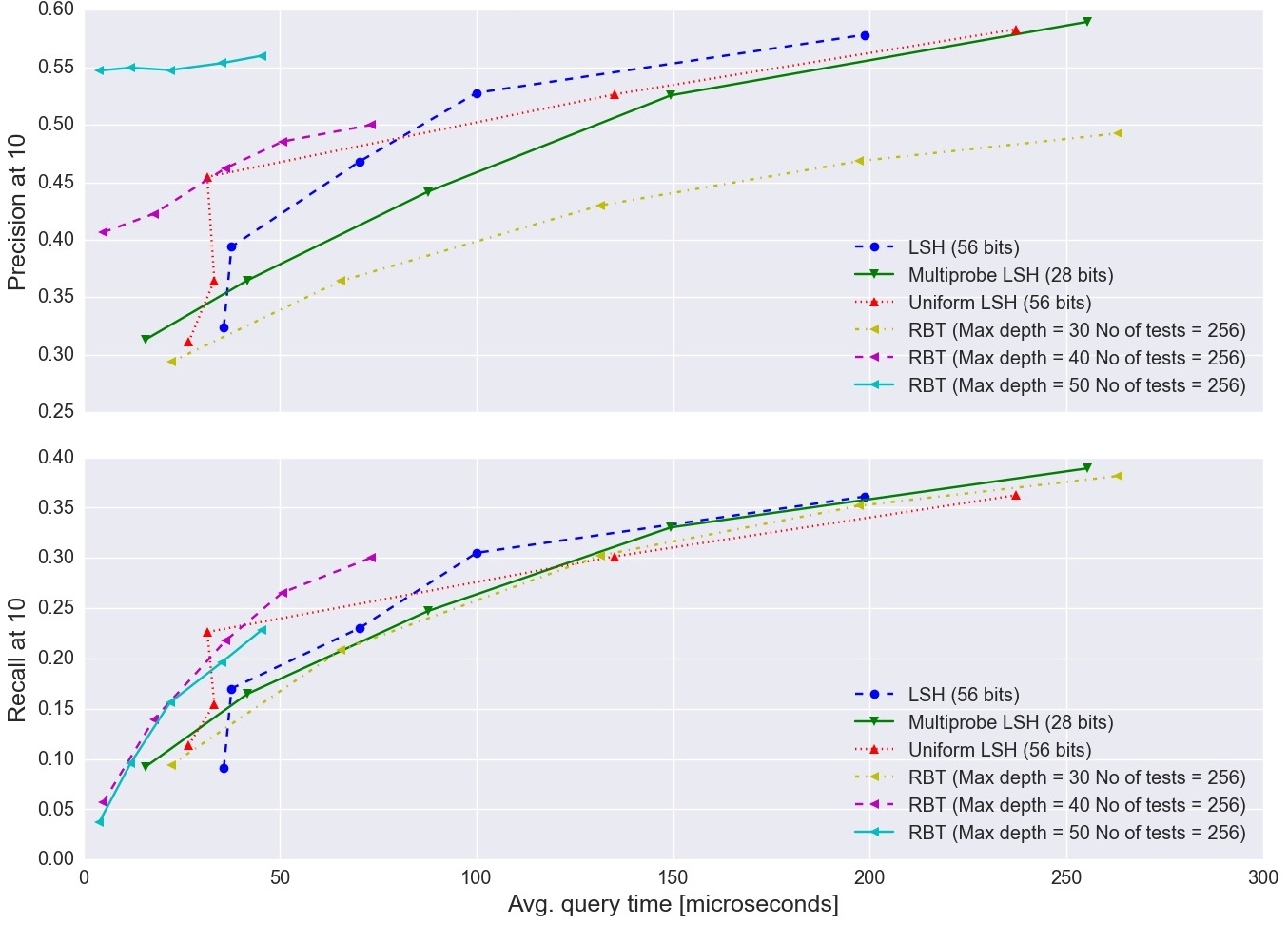}
    \caption{\prten{} and \recallten{} versus the \aqt{}.}
    \label{fig:benchmark_release_build}
\end{figure}

\begin{figure}[t]
    \centering
    \includegraphics[width=\textwidth,height=7cm]{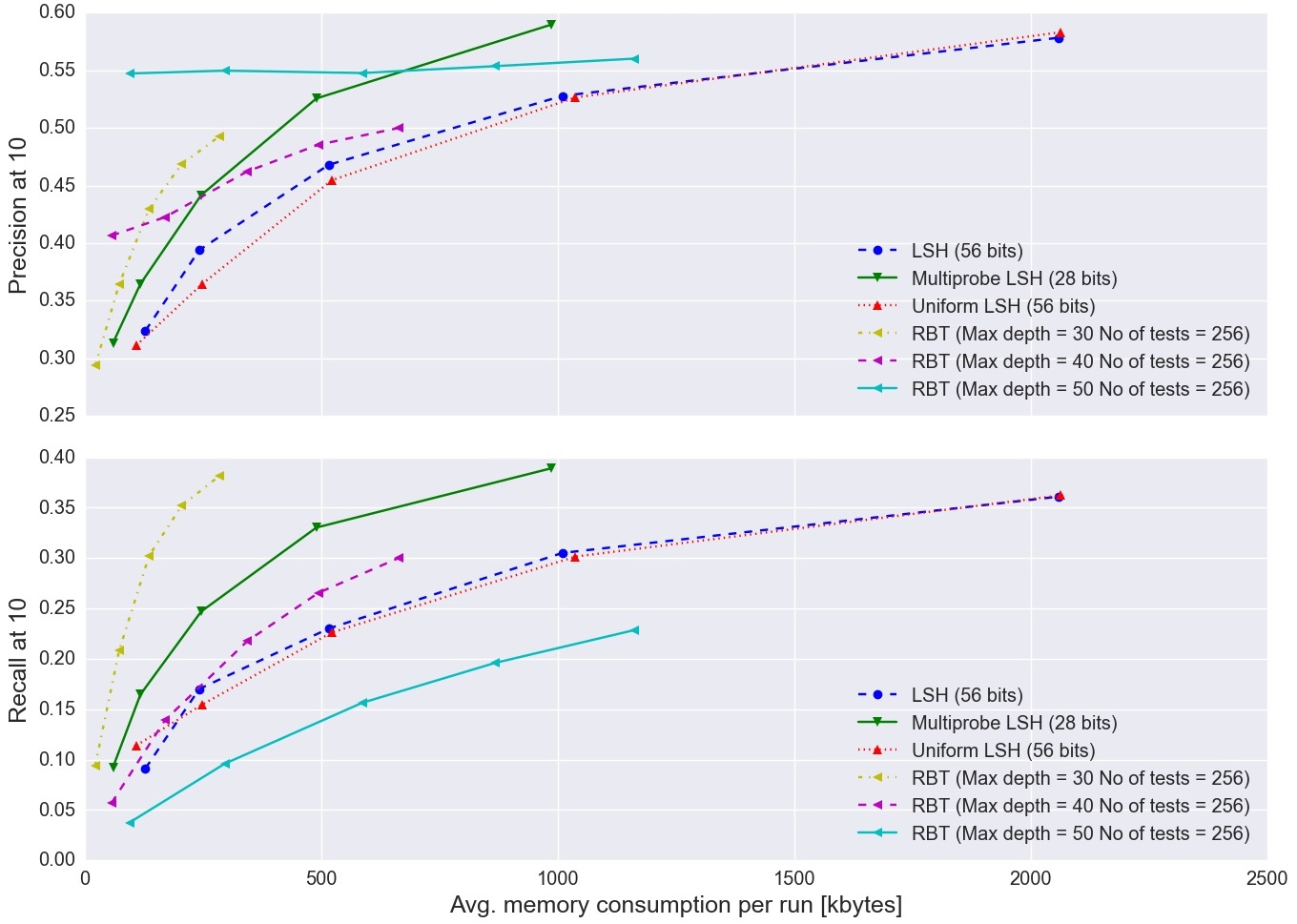}
    \caption{\prten{} and \recallten{} versus the \amc{}.} 
    \label{fig:benchmark_memory_release_build}
\end{figure}

In this section, we compare our \rbt{} against the competitive approaches for ANN search in binary spaces. Figures \ref{fig:benchmark_release_build} and \ref{fig:benchmark_memory_release_build} show the results of experiments. Following the evaluation protocol of~\cite{trzcinski2012thick} we plot Precision and Recall results obtained against average query times. We compare our method against 3 variants of Local Sensitive Hashing (LSH) algorithm, as they were shown to provide the best performances in~\cite{trzcinski2012thick}. We use our own implementation of those algorithms. The parameters of all the methods were optimised using grid search approach. In the case of \rbt{}  the evaluation was done for  \notr{}~$ = \{1, 3, 6, 9, 12\}$ trees. For the hashing methods, the number of hash tables used were equal to $\{1, 2, 4, 8, 16\}$. For LSH and Uniform LSH the hash length was $56$ and for Multi-Probe $28$. We report \amc{} as memory required by the algorithms to build indexing structures for descriptors and not descriptors themselves. 

Figure \ref{fig:benchmark_release_build} shows that \rbt{} provides better performance with respect to the state of the art hashing methods in terms of search precision, given equal query time. The performance boost is especially visible for lower average query times ($<$50 $\mu$seconds), where our proposed \rbt{} algorithm leads to over 20\% precision increase over the next best Uniform-LSH method. At the same time, our evaluation shows that the precision increase does not lead to any significant recall drops. If we consider both Precision and Recall our \rbt{} achieve the best results for \md{} = $40$ and \mnot{}$ = 256$. 

Figure \ref{fig:benchmark_memory_release_build} compares various methods in terms of memory consumption. Although particular results depend on the tested configuration, one can see that for \md{} = $40$ and \mnot{}$ = 256$ \rbt{} performs au pair with the state-of-the-art methods, falling short only of the Multi-probe LSH, which is highly optimised for memory consumption. We can therefore conclude that our proposed \rbt{}  search method provides significant precision increase, while remaining competitive in terms of recall and memory consumption.

\section{Summary}

In this article, we proposed to use \rbtfull{} (\rbt{}) algorithm to index and search binary descriptors. We tested a wide range of configurations and we compared them with Locality Sensitive Hashing (LSH) and its two variations. The experiments showed that, although \rbt{} are a relatively simple data structure, they give better or equal results to the competing  hashing algorithms.

Future work on ANN search with our trees includes improving the linear search stage after retrieving the initial set of candidate descriptors, as this part remains a bottleneck of the algorithm. Furthermore, application of a more complex bit metric that can measure dependencies between the bits could lead to the improved precision and search efficiency and should also remain within the scope of future work.  


\section*{Acknowledgment}
This research was supported by Google's Sponsor Research Agreement under the project "Efficient visual localisation on mobile devices".

\end{document}